\documentclass[conference]{IEEEtran}
\IEEEoverridecommandlockouts
\usepackage{cite}
\usepackage{amsmath,amssymb,amsfonts}
\allowdisplaybreaks
\usepackage{algorithmic}
\usepackage{graphicx}
\usepackage{textcomp}
\usepackage{xcolor}
\usepackage{multicol,multirow}
\usepackage{marvosym}
\def\BibTeX{{\rm B\kern-.05em{\sc i\kern-.025em b}\kern-.08em
    T\kern-.1667em\lower.7ex\hbox{E}\kern-.125emX}}

\newcommand{\textred}{\textcolor{red}}
\newcommand{\textblue}{\textcolor{blue}}

\IEEEpubid{\makebox[\columnwidth]{979-8-3503-3748-8/23/\$31.00~\copyright~2023 IEEE\hfill}%
\hspace{\columnsep}\makebox[\columnwidth]{ }}

\begin{document}

\title{
ScribblePolyp: Scribble-Supervised Polyp Segmentation through Dual Consistency Alignment
}

\author{
    \IEEEauthorblockN{1\textsuperscript{st} Zixun Zhang$\star$\thanks{$\star$~Equal contributions.}}
    \IEEEauthorblockA{
        \textit{FNii, CUHK-Shenzhen} \\
        \textit{SSE, CUHK-Shenzhen}\\
        Shenzhen, China \\
        zixunzhang@link.cuhk.edu.cn
    }
    \and
    \IEEEauthorblockN{2\textsuperscript{nd} Yuncheng Jiang$\star$, }
    \IEEEauthorblockA{
        \textit{FNii, CUHK-Shenzhen} \\
        \textit{SSE, CUHK-Shenzhen}\\
        Shenzhen, China \\
        yunchengjiang@link.cuhk.edu.cn
    }
    \and
    \IEEEauthorblockN{3\textsuperscript{rd} Jun Wei}
    \IEEEauthorblockA{
        \textit{FNii, CUHK-Shenzhen} \\
        \textit{SSE, CUHK-Shenzhen}\\
        Shenzhen, China \\
        junwei@link.cuhk.edu.cn
    }
    \and
    \IEEEauthorblockN{4\textsuperscript{rd} Hannah Cui}
    \IEEEauthorblockA{
        \textit{The Chinese University of Hong Kong, Shenzhen} \\
        Shenzhen, China \\
        hannah.cui6@gmail.com
    }
    \and
    \IEEEauthorblockN{5\textsuperscript{th} Zhen Li\textsuperscript{\Letter}\thanks{\textsuperscript{\Letter}~Corresponding Author}}
    \IEEEauthorblockA{
        \textit{SSE, CUHK-Shenzhen}\\
        \textit{FNii, CUHK-Shenzhen} \\
        Shenzhen, China \\
        lizhen@cuhk.edu.cn
    }
}

\maketitle

\begin{abstract}

Automatic polyp segmentation models play a pivotal role in the clinical diagnosis of gastrointestinal diseases. 
In previous studies, most methods relied on fully supervised approaches, necessitating pixel-level annotations for model training. 
However, the creation of pixel-level annotations is both expensive and time-consuming, impeding the development of model generalization. 
In response to this challenge, we introduce ScribblePolyp, a novel scribble-supervised polyp segmentation framework. 
Unlike fully-supervised models, ScribblePolyp only requires the annotation of two lines (scribble labels) for each image, significantly reducing the labeling cost. 
Despite the coarse nature of scribble labels, which leave a substantial portion of pixels unlabeled, we propose a two-branch consistency alignment approach to provide supervision for these unlabeled pixels. 
The first branch employs transformation consistency alignment to narrow the gap between predictions under different transformations of the same input image. 
The second branch leverages affinity propagation to refine predictions into a soft version, extending additional supervision to unlabeled pixels. 
In summary, ScribblePolyp is an efficient model that does not rely on teacher models or moving average pseudo labels during training. 
Extensive experiments on the SUN-SEG dataset underscore the effectiveness of ScribblePolyp, achieving a Dice score of 0.8155, with the potential for a 1.8\% improvement in the Dice score through a straightforward self-training strategy.
\end{abstract}


\section{Introduction}

Colonoscopy is a commonly used method for checking intestinal diseases, and early detection of polyps is an effective means for preventing colon cancer. 
Therefore, polyp segmentation serves as an important task in the colonoscopic image analysis: it can help doctors to diagnose and treat polyps more accurately and efficiently. 
Recently, with the development of deep learning, many polyp segmentation methods~\cite{sanet,polyppvt} have been proposed and made great progress. 
However, these methods are data hungry, which require a large amount of fully annotated mask labels. 
Unfortunately, labeling polyp masks is time-consuming and laborious, requiring the guidance of professional doctors.

\begin{figure}[t]
    \centering
    \includegraphics[width=0.9\linewidth]{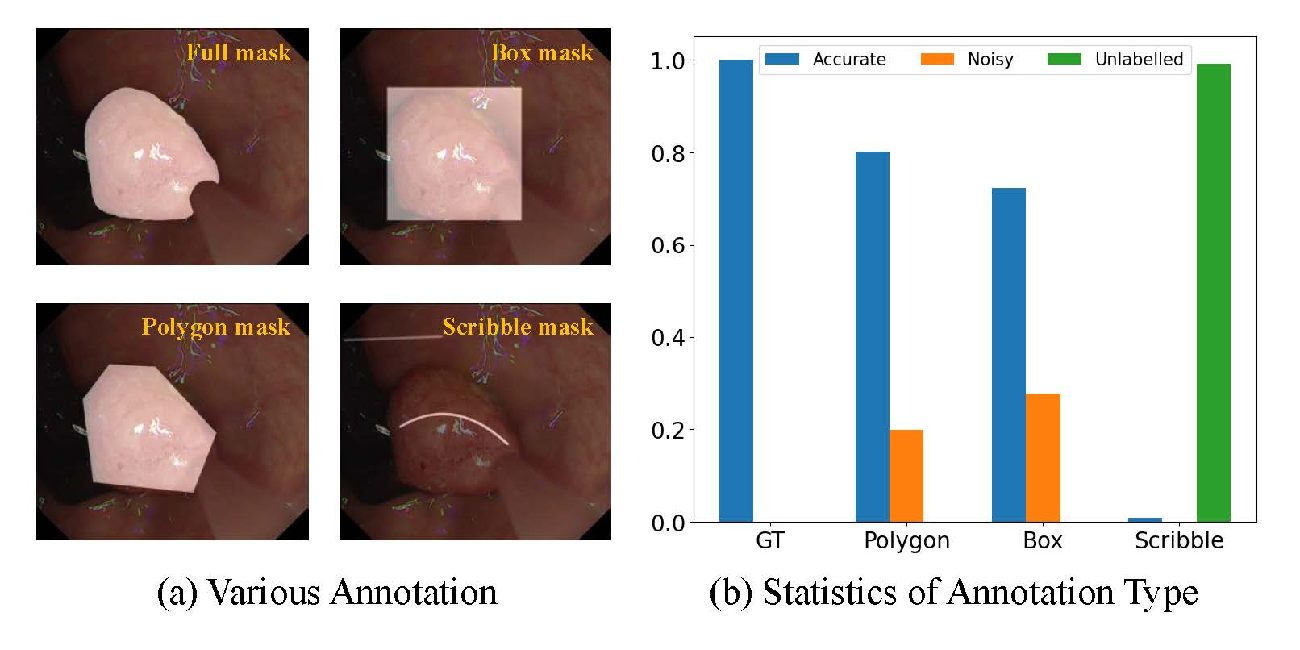}
    \caption{
    (a) Example annotations of full mask, box mask, polygon mask and scribble mask. (b) Accuracy of different types of annotations. 
    }
    \label{fig:stat}
\end{figure}

Recent studies in medical image analysis have begun to focus on weakly supervised segmentation.
Fig.~\ref{fig:stat}(a) shows four forms of weak annotation (i.e., box, polygon, scribble, point) and Fig.~\ref{fig:stat}(b) counts the proportion of pixels with accurate labels, noisy labels, and no labels in each type of weak annotation.
From Fig.~\ref{fig:stat}(b), box and polygon annotations contain a considerable amount of accurate labels (over 70\%). These labels have correct annotations and can be used as supervision signals.
However, the annotation correctness of remaining pixels, taking about 20\% $\sim$ 30\%, cannot be guaranteed.
For scribble labels, the situation is completely different. Only a small amount of pixels have accurate labels (less than 10\%), while the rest of the pixels are unlabeled. 

In scribble-supervised segmentation, methods to supervise unlabeled regions, like generating pseudo-labels or self-supervised learning, are essential. Previous approaches maintain pseudo labels throughout training, such as using moving average predictions or teacher models to generate these pseudo labels~\cite{scribble2label,ustm}.
However, these methods are time-consuming as they require frequent updates of pseudo labels or teacher models. Furthermore, maintaining pseudo labels can accumulate inaccurate annotations, potentially misleading model training.

To address these concerns, we introduce ScribblePolyp for scribble-supervised polyp segmentation. 
It employs a two-branch consistency alignment approach to supervise unlabeled regions.
In the first branch, we implement transformation consistency alignment to ensure consistency between predictions under different transformations. 
Specifically, we apply resizing and cropping transformations to align predictions, focusing on scale consistency and local-global coherence. 
However, this alignment alone doesn't guide the model in distinguishing foreground and background. 
For the second branch, we propose an affinity propagation (AP) alignment to connect labeled and unlabeled pixels, providing additional supervision. 
Inspired to non-local~\cite{nonlocal} and self-attention~\cite{vit} mechanisms, we create a reliable similarity matrix between labeled and unlabeled pixels. 
We use encoder features to generate a pixel-level affinity map and propagate the original prediction to produce a soft prediction. 
This soft prediction is then aligned with the original prediction, capturing extra supervision from affinity. 
Additionally, we incorporate a simple self-training step in ScribblePolyp for performance enhancement, avoiding the accumulation of inaccurate pseudo annotations and saving computational resources.

In summary, our contributions are three-folds:
(1) We introduce ScribblePolyp, a novel training framework with dual consistency alignment for efficient scribble-supervised polyp segmentation, substantially reducing labeling costs.
(2) To address the challenge of unlabeled regions, we propose the transformation consistency alignment and affinity propagation alignment, ensuring prediction consistency under various transformations and leveraging pixel similarity for enhanced supervision.
(3) Extensive experiments on the SUN-SEG dataset~\cite{sun1,sun2,sunseg} demonstrate the effectiveness of our proposed ScribblePolyp. 
In particular, it achieves a Dice score of \textbf{0.8302}, surpassing the partial binary cross entropy (BCE) baseline by \textbf{13.9\%} and trailing the fully-supervised SANet by only \textbf{4.6\%}.

\begin{figure}[t]
    \centering
    \includegraphics[width=0.95\linewidth]{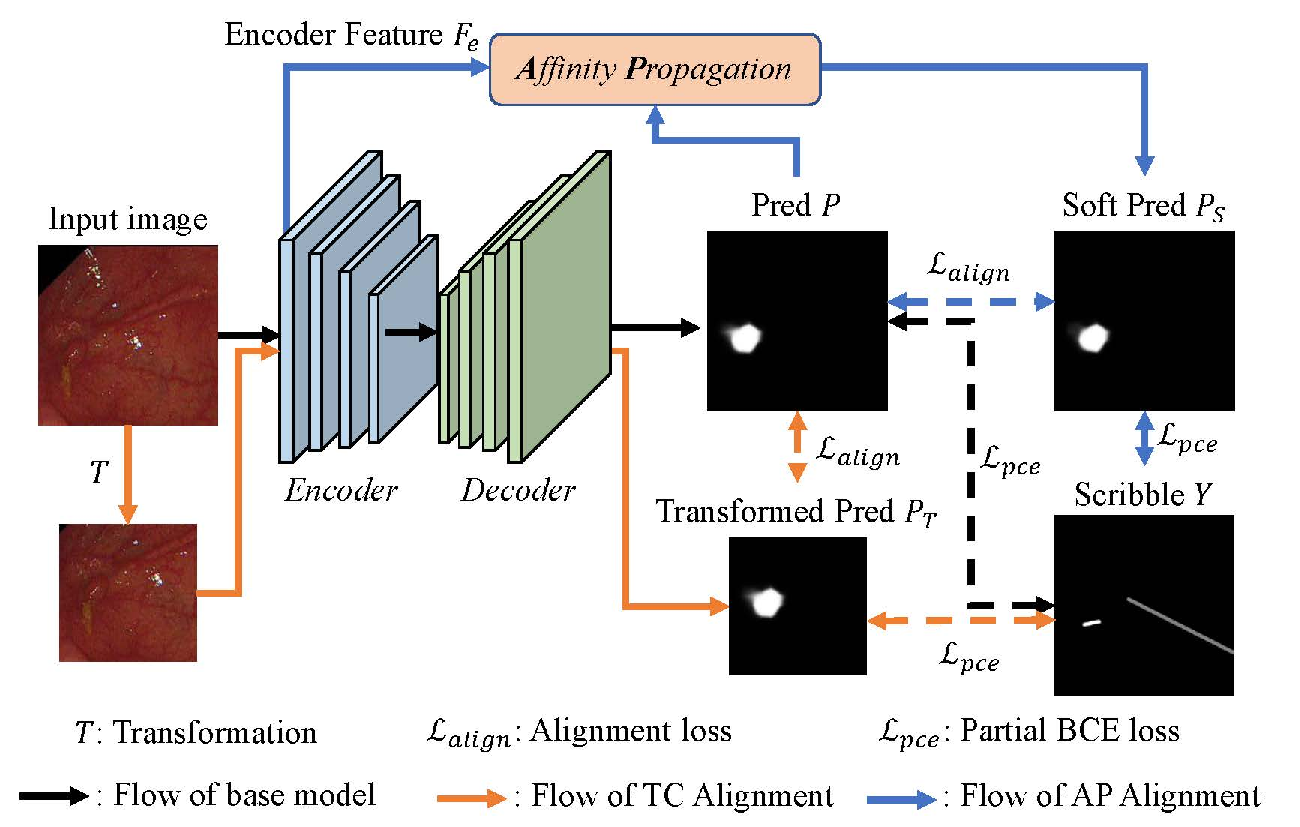}
    \caption{Our proposed ScribblePolyp for scribble-supervised polyp segmentation.
    In the \textcolor{orange}{orange} branch, our ScribblePolyp performs the transformation consistency alignment to align the predictions of the same input image to reduce the inconsistency.
    While in the \textcolor{blue}{blue} branch, our ScribblePolyp performs the affinity propagation alignment which leverages the affinity map to refine the original prediction to bridge the labeled and unlabeled pixels.
    }
    \label{fig:framework}
\end{figure}

\section{Method}

The overall learning framework of ScribblePolyp is illustrated in Fig.~\ref{fig:framework}, which contains a base segmentation model and two additional alignment modules. 
For pixels with scribble labels, we train the corresponding pixels of the prediction with partial BCE loss $\mathcal{L}_{pce}$.

\begin{align*}
    \mathcal{L}_{pce} = -\frac{1}{ \left | S \right | } \sum_{i \in S} (Y_i\log{P_i}+(1-Y_i)\log{(1-P_i)})
\end{align*}

\noindent where $Y_i$ and $P_i$ denote the label and prediction at location $i$, and $S$ denotes the set of pixels annotated with scribble labels. 
For unlabeled pixels, ScribblePolyp introduces the two-branch consistency alignments for supervision.

\subsection{Transformation Consistency Alignment}
\label{sec:tca}

As mentioned above, for scribble-supervised polyp segmentation, only a small number of pixels can be supervised during training.
Therefore, the segmentation model tends to suffer from inconsistent predictions.
\cite{dmpls} proposed to add an auxiliary decoder which takes dropout encoder features as input and mixes the predictions as pseudo labels to reduce inconsistency.
Another study~\cite{ustm} proposed to leverage the inconsistency for measuring the uncertainty to exclude the unreliable pixels predicted by mean teacher model.
In this paper, we propose to reduce the inconsistency of predictions under different transformations of the same input image.
In fully supervised polyp segmentation, the transformation consistency is guaranteed by the same augmentation strategy of both images and labels, which introduces an implicit equivariant constraint for the network.
While in scribble-supervised polyp segmentation, this implicit constraint is not guaranteed due to the lack of supervision.

In particular, we perform two types of transformations (i.e., scale consistency and local-global consistency) to the input image and then align the original prediction $P$ and the transformed prediction.

\subsubsection{Scale Consistency Alignment} 
In scale consistency (sc), given the original input image $I$ with resolution $H \times W$, we randomly resize the input images to another resolution $H_{sc} \times W_{sc}$. Then we resize the transformed prediction $P_{sc}$ back to the original resolution, since the test is performed on original resolution. Finally, we reduce the distance between $P$ and the resized $P_{sc}$ by MSE loss.

\begin{align*}
    \mathcal{L}_{sc} &= \frac{\sum_{H} \sum_{W} (P - P_{sc, r})^2}{H \times W}
\end{align*}

Since the predictions with different scale transformations are equivalent to each other from the global view, $\mathcal{L}_{sc}$ does not need to detach the gradient of the original prediction and directly lets both predictions get closer instead.

\subsubsection{Local-global Consistency Alignment} 
Although scale consistency alignment reduces the inconsistency of predictions between different scales, such alignment is performed on the global view, ignoring the local details.
Therefore, in addition to the scale consistency, we propose an alignment for the local and global views to improve the consistency of local details.
Specifically, in the local-global consistency (lg) alignment, we randomly crop a local patch from the original input image to get the local prediction $P_{l}$. 
Then we crop the corresponding patch from the original prediction $P$ to obtain the global prediction $P_{g}$. 
Finally, we align the local and global prediction via MSE loss similar to the sc alignment.

\subsection{Affinity Propagation Alignment}
\label{sec:apa}

In transformation consistency alignment, we ameliorate the inconsistency issue by reducing distance between predictions of different types of transformations. 
However, such an alignment could only guarantee the consistency of the prediction of the same input sample, but the relationships between pixels are ignored.
Inspired by this, we propose to build the affinity map between pixels and use this map to propagate supervision signals from labeled pixels to unlabeled ones.
Similar to non-local~\cite{nonlocal} or self-attention~\cite{vit}, we extract contexts from encoder to build the affinity map.
Through affinity propagation, unlabeled pixels get more supervision, thereby the decision boundaries of the model become precise.

\begin{figure}[t]
    \centering
    \includegraphics[width=0.9\linewidth]{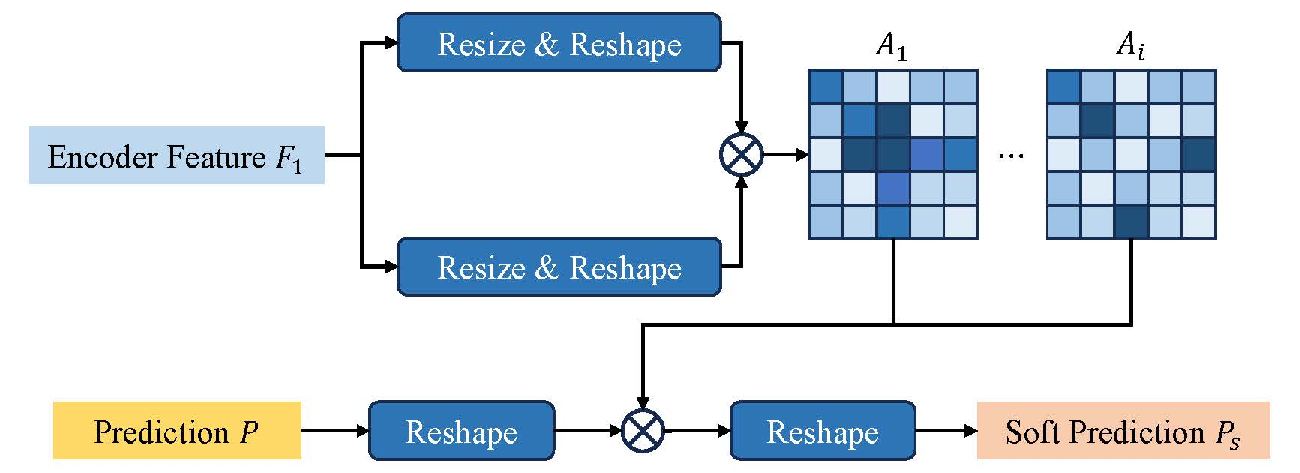}
    \caption{Procedure of multi-level affinity propagation. The softmax operation is omitted in the figure.}
    \label{fig:affinity_propagation}
\end{figure}

\subsubsection{Affinity Propagation}

Given the input image $I$, the segmentation model first produces a prediction $P$, and a series of encoder features with different scales, e.g., $\times 4$, $\times 8$, $\times 16$, $\times 32$ from common CNN backbones. 
Thus, we can take the encoder features $F_e$ to produce the affinity map for each pixels.
Specifically, we first resize the features to the size of prediction and reshape them to $\mathbb{R}^{B \times L \times C}$ as embeddings $E$, where $L=H \times W$ and $B$, $C$ are batch size and channels, respectively. 
Then these embeddings are multiplied with each other to produce the affinity map $A \in \mathbb{R}^{HW \times HW}$ and normalized by softmax operation.
This affinity map contains a pairwise similarity for each pixel $i$ and $j$, including the relationship between labeled pixels and unlabeled pixels.
Finally, this affinity map is used to refine the original prediction $P$ to obtain the soft prediction $P_s$.

\subsubsection{Multi-level Affinity Propagation}

Since the encoder can produce multiple features with different scales, these features contain different contextual information.
Here, we introduce a multi-level affinity propagation to propagate the prediction multiple times with features of different scales.
Specifically, given the affinity maps $\{A_{i}\}$ produced by different levels of encoder features, we successively multiply these affinity maps with the original prediction to obtain the soft prediction $P_s$, as shown in Fig.~\ref{fig:affinity_propagation}.

Finally, the original prediction is aligned with this soft prediction by $\mathcal{L}_{ap}$ using MSE. 
Such multiple propagations can take full advantage of multi-scale features, thus, the model could further obtain more precise supervision from affinity map to learn the global contextual information outside scribble labels.

\subsection{Overall Training Pipeline}
\label{sec:training}

During the training process, we randomly select one consistency alignment from transformation consistency alignment and affinity propagation alignment to supervise the segmentation model along with the partial BCE loss. 
We empirically find this training strategy is more efficient, and the performance is almost the same as training with both alignments.
The total training loss is as follow:

\begin{align*}
    \mathcal{L}_{tot} &= \mathcal{L}_{pce} + \mathcal{L}_{align} \\
    \mathcal{L}_{align} &\in \{\mathcal{L}_{sc}, \mathcal{L}_{lg}, \mathcal{L}_{ap}\}
\end{align*}

Additionally, the simple self-training could further boost the performance of ScribblePolyp.
After obtaining a well-trained segmentation model with dual consistency alignments, we take it as the teacher model to generate the pseudo labels.
Compared with EMA-based methods~\cite{ustm,scribble2label}, these labels are more reliable since they make full use of the contextual information extracted by the encoder, and there is no accumulation of wrong labels like EMA-based methods. 
Moreover, this one-time generation approach is more efficient. 
Compared with maintaining a moving average of pseudo labels or teacher model, the computational cost of our pseudo-label generation method is negligible.
More than that, without repeated pseudo-label generation, the overall training time of ScribblePolyp is much shorter than that of the EMA-based methods~\cite{ustm,scribble2label}.
With these generated pseudo labels, we further train a basic segmentation model without incorporating consistency alignment or affinity propagation.
Afterwards, we use this basic segmentation model as the final model for inference and evaluation.

\section{Experiment}

\subsection{Dataset and Implementation Details}

To verify the performance of our proposed ScribblePolyp, we evaluate our ScribblePolyp on SUN-SEG dataset~\cite{sunseg} which is based on SUN Database~\cite{sun1,sun2}.
SUN-SEG contains 285 video clips from 100 distinct video polyp cases. 
To reduce the data redundancy of video frames, we follow the data split used in \cite{s2me}, which using 6677 for training, 1240 for validation and 1993 for testing.

Here we choose SANet~\cite{sanet} as our base polyp segmentation model and add the proposed alignment losses for training. 
It is worth noting that ScribblePolyp is general and can be combined with most polyp segmentation models.
The augmentation is following the settings in SANet.
The model is trained by SGD with momentum of 0.9 with a initial learning rate of $1e-3$ for 64 epochs.
We compare our method with 3 scribble-supervised methods, Scribble2Label~\cite{scribble2label}, DMPLS~\cite{dmpls} and USTM~\cite{ustm}, and re-implement their codes as the same settings for fair comparison.

\subsection{Comparison with Other Methods}

The results are shown in Tab.~\ref{fig:compare}. We compare ScribblePolyp with three state-of-the-art methods for scribble-supervised medical image segmentation.
Our ScribblePolyp achieves the best Dice and IoU scores, which demonstrate the superior performance of our proposed ScribblePolyp framework.
Specifically, our ScribblePolyp achieves a Dice score of 0.8155 and a IoU score of 0.7200, surpassing the second best method Scribble2Label~\cite{scribble2label} by 2.3\% and 3.2\%, respectively.
At the same time, our ScribblePolyp also achieves the best precision score of 0.8195, which surpasses the second best result by 7.4\%.
Although the recall of the DMPLS~\cite{dmpls} and partial BCE baseline exceeds 0.9, their precision results indicate that they segment many background regions as polyp targets, which leads to an imbalance between precision and recall.
The same phenomenon occurs in other methods.
In contrast, the precision and recall of our ScribblePolyp are relatively balanced.
Compared with the partial BCE baseline, our ScribblePolyp achieves an improvement of 11.9\% on Dice score and 16.8\% on IoU score, which demonstrates the superior learning ability of our method.
Furthermore, with a simple self-training, our ScribblePolyp achieves an improvement of 1.8\% on Dice score and 2.6\% on IoU score. 
Compared with the fully supervised SANet~\cite{sanet}, our ScribblePolyp is only 4.6\% lower on Dice score and 8.1\% lower on IoU score.
All the results demonstrate our good capability for scribble-supervised polyp segmentation.

\begin{table}[t]
	\centering
	\caption{
		Results of our ScribblePolyp and comparison methods. 
        Sup. denotes the supervision labels used for training.
        F denotes full mask, S denotes scribble and P denotes pseudo label.
		$\dagger$: we remove the mixed pseudo labels of DMPLS~\cite{dmpls}.
		The best score is marked as \textred{red}, while the second best score is marked as \textblue{blue}.
	}
	\begin{tabular}{r|c|ccccc}
		\hline
		Method & Sup. & Dice & IoU & Precision & Recall\\
		\hline
		SANet~\cite{sanet} & F & 0.8705 & 0.8039 & 0.8841 & 0.8850 \\
		\hline
		Partial BCE & S & 0.7284 & 0.6165 & 0.6627 & \textblue{0.9037} \\
		DMPLS$\dagger$~\cite{dmpls} & S & 0.6636 & 0.5358 & 0.5721 & \textred{0.9058} \\
		DMPLS~\cite{dmpls} & S+P & 0.4454 & 0.3163 & 0.3299 & 0.9025 \\
		USTM~\cite{ustm} & S+P & 0.7480 & 0.6361 & 0.6955 & 0.8935 \\
		Scribble2Label~\cite{scribble2label} & S+P & 0.7968 & 0.6976 & 0.7631 & 0.8943 \\
		\hline
		ScribblePolyp (ours) & S & \textblue{0.8155} & \textblue{0.7200} & \textblue{0.8195} & 0.8581 \\
		~+ self-train (ours) & S+P & \textred{0.8302} & \textred{0.7390} & \textred{0.8334} & 0.8684 \\
		\hline
	\end{tabular}
	\label{fig:compare}
\end{table}

\section{Conclusion}

Considering that obtaining full mask annotations of polyp requires a lot of time and expertise, in this paper, we propose the ScribblePolyp, a powerful learning framework for scribble-supervised polyp segmentation.
ScribblePolyp leverages a dual consistency alignment to provide supervision for unlabeled regions. 
In particular, ScribblePolyp utilizes a transformation consistency alignment to reduce the inconsistency of predictions under different transformations, and utilizes affinity map generated by encoder features to propagate the prediction, bridging the relationship between labeled and unlabeled regions.
With a simple self-training, our ScribblePolyp could further obtain a performance improvement.
Extensive experiments and ablations verify the effectiveness of our method.

\section*{Acknowledgements}

This work was supported in part by Shenzhen General Program No.JCYJ20220530143600001, by the Basic Research Project No.HZQB-KCZYZ-2021067 of Hetao Shenzhen HK S\&T Cooperation Zone, by ShenzhenHong Kong Joint Funding No.SGDX20211123112401002, by Shenzhen Outstanding Talents Training Fund, by Guangdong Research Project No.2017ZT07X152 and No.2019CX01X104, by the Guangdong Provincial Key Laboratory of Future Networks of Intelligence (Grant No.2022B1212010001), The Chinese University of Hong Kong, Shenzhen, by the NSFC 61931024\&81922046, by Tencent Open Fund.

\bibliographystyle{IEEEtran}
\bibliography{IEEEabrv,bibtex}

\end{document}